\documentclass[letterpaper,10pt,conference]{ieeeconf}  
\IEEEoverridecommandlockouts
\usepackage{cite}
\usepackage{amsmath,amssymb,amsfonts}
\usepackage{algorithmic}
\usepackage{graphicx}
\usepackage{textcomp}
\usepackage{xcolor}
\usepackage{booktabs}
\usepackage[most]{tcolorbox}
\usepackage{url}
\usepackage{hyperref}

\begin{document}

\title{Interpretable Language Model for Closed-Loop Type 1 Diabetes Control }

\author{\authorblockN{Maya Sarkar}
\authorblockA{\textit{Visaze LLC, California, USA} \\
mayasarkar010@gmail.com}}

\maketitle
\begin{abstract}

Type 1 Diabetes (T1D) is a chronic, life-threatening autoimmune condition characterized by the complete destruction of insulin-producing pancreatic beta cells. While Artificial Pancreas Systems (APS) powered by Reinforcement Learning (RL) have shown promise in automating insulin delivery, their ``black-box'' nature makes it hard for patients and doctors to trust them fully. This paper presents LLM-T1D, a promising approach that combines the precision of RL with the clear, human-like reasoning of Large Language Models (LLMs) to create a more transparent and reliable insulin pump controller. By training an expert RL system and distilling its knowledge into fine-tuned LLaMA 3.1 8B and Qwen3 8B models, we developed a controller that not only surpasses the RL system's performance but also explains its decisions in plain, understandable language. Tested on the FDA-approved UVA/Padova T1D simulator, the LLM controllers deliver excellent blood sugar control (73.5\% Time in Range) while maintaining strict formal safety verification against hallucinations.
\end{abstract}

\noindent\textbf{Demo Video}: \url{https://youtu.be/ljzBz938Gbo}

\section{Introduction}

Type 1 Diabetes (T1D) is a chronic autoimmune disease in which the loss of insulin-producing pancreatic beta cells prevents native blood-glucose regulation, requiring lifelong exogenous insulin therapy \cite{hettiarachchi2024g2p2c}. For patients, daily management is a continuous control problem: maximize Time in Range (TIR), typically 70--180 mg/dL \cite{gabbay2020time}, while avoiding acute hypoglycemia and long-term hyperglycemic complications \cite{dimeglio2018type}.

Artificial Pancreas Systems (APS) seek to reduce this burden through automated insulin delivery \cite{breton2021one}. Classical PID and MPC controllers have been useful for basal regulation, but they remain limited by nonlinear insulin dynamics, meal variability, and large inter-patient differences. Reinforcement Learning (RL) provides a more adaptive alternative by modeling glucose regulation as a partially observable sequential decision problem \cite{hettiarachchi2024g2p2c}. Prior work has shown that PPO-based methods can learn personalized dosing policies from complex physiological state information and may reduce reliance on manual carbohydrate announcements \cite{fox2020deep}.

\begin{figure}[t]
\centering
\includegraphics[width=\columnwidth]{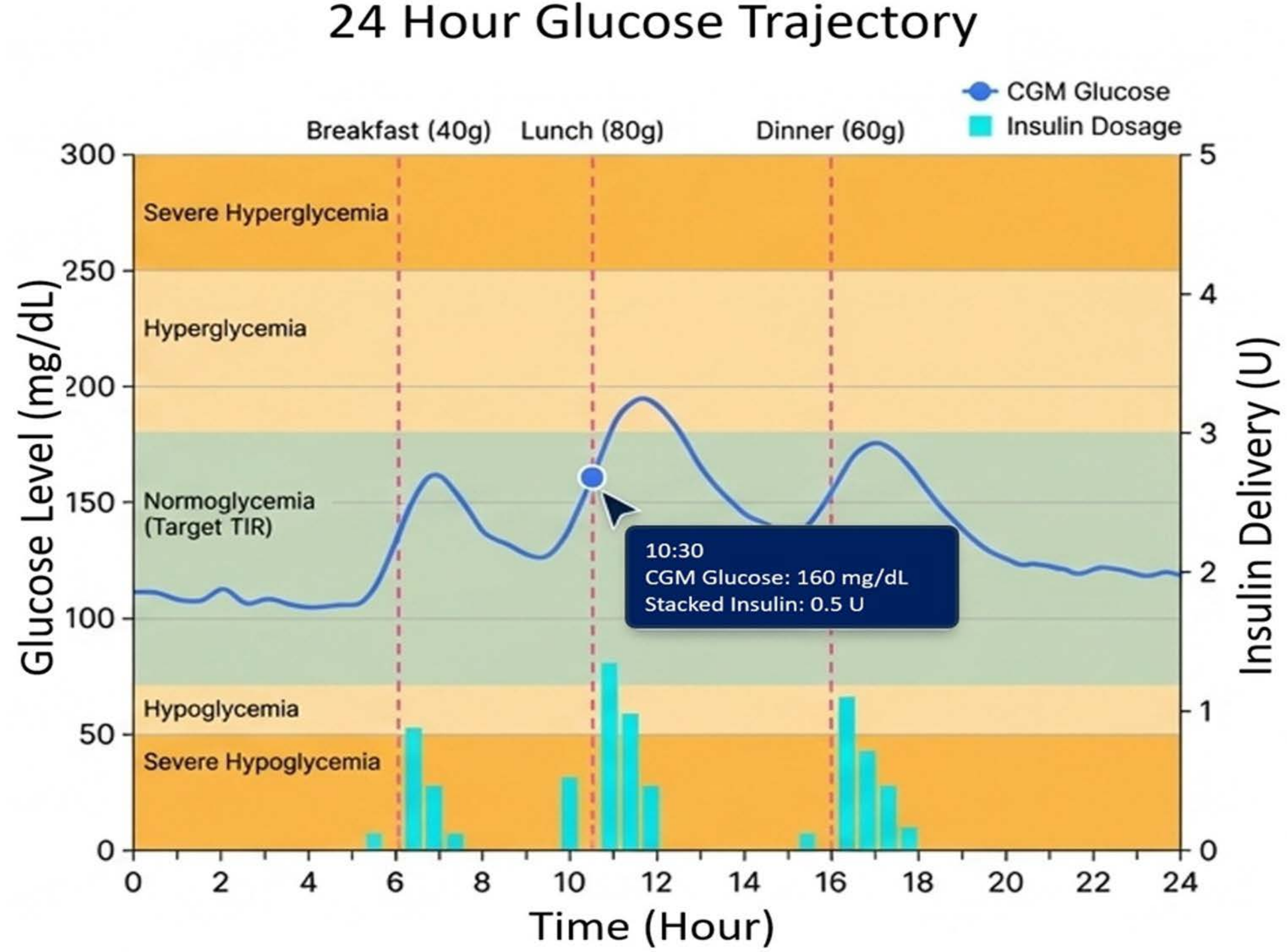}
\caption{The autonomous LLM-T1D controller effectively manages continuous glucose fluctuations over a highly demanding 24-hour period. Despite facing three completely unannounced meals (40g, 80g, and 60g CHO), the semantic policy successfully predicts the required insulin compensation, mitigating extreme hyperglycemic spikes and entirely preventing hypoglycemic crashes below the 70 mg/dL boundary.}
\label{fig:daily}
    \vskip -0.4 cm
\end{figure}

\begin{figure}[t]
    \centering
    \includegraphics[width=1\columnwidth]{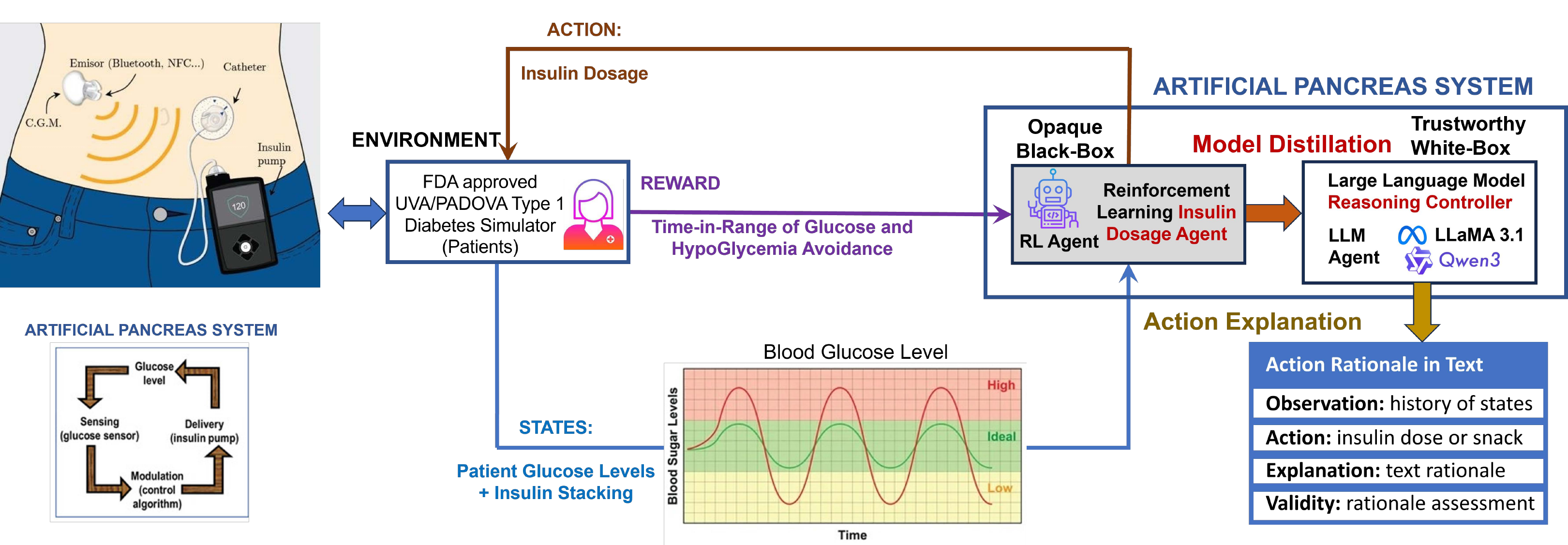}
    \caption{System Diagram for LLM based insulin pump control}
    \label{fig:sys}
    \vskip -0.4 cm
\end{figure}

\begin{figure}[h]
\centering
\includegraphics[width=\columnwidth]{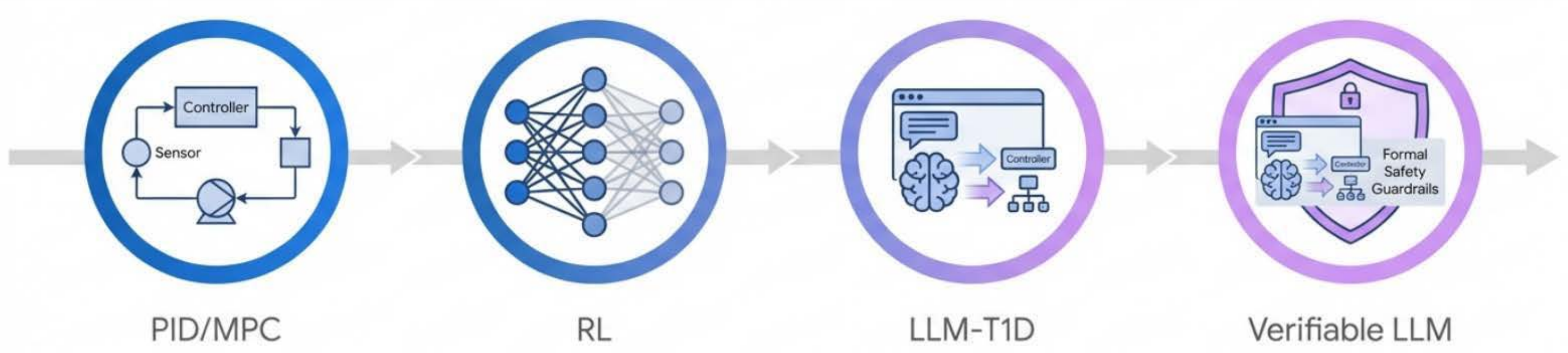}
\caption{The Evolution of Artificial Pancreas Control Architectures. The progression of Artificial Pancreas Systems (APS) highlights the critical transition from opaque mathematical optimization (Reinforcement Learning) to transparent, verifiable, and semantic control (LLM-Distillation).}
\label{fig:evolution}
    \vskip -0.4 cm
\end{figure}
A key barrier, however, is clinical trust. RL-based APS controllers can perform well in \textit{in-silico} studies, but their dosing decisions are often difficult for patients and clinicians to interpret \cite{lee2021bioinspired}. A neural policy may recommend insulin without explaining the role of glucose trend, insulin-on-board, meal context, or patient-specific sensitivity \cite{gokhale2024explainable, liu2025beyond}. In a safety-critical medical device, this opacity creates practical concerns around oversight, liability, and patient acceptance \cite{gokhale2024explainable, lim2021blood}, especially when an incorrect recommendation could cause severe hypoglycemia \cite{chang2024security, flory2025llm}.

\begin{figure*}[t]
    \centering
    \includegraphics[width=2\columnwidth]{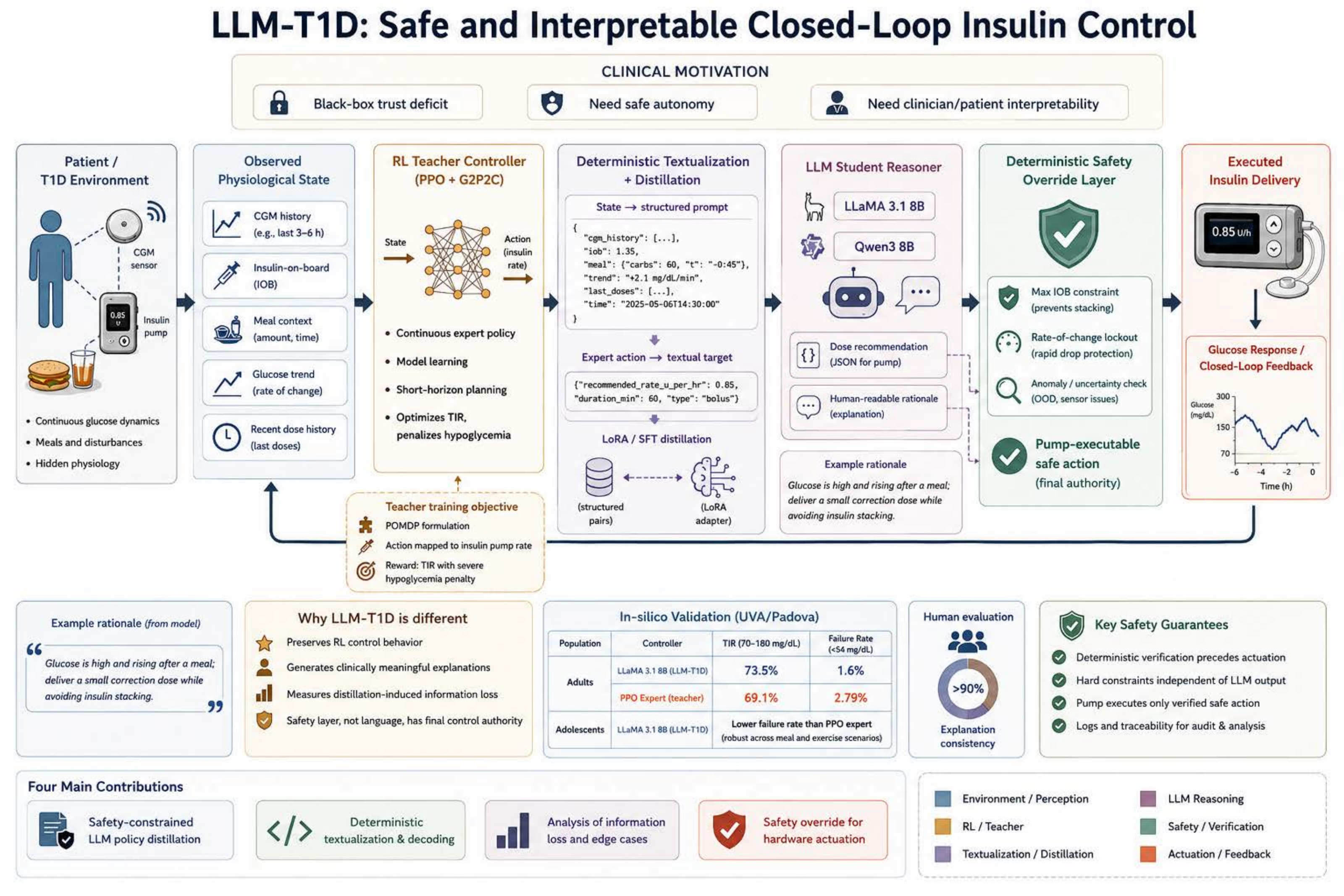}
    \caption{LLM-T1D distills an expert RL insulin controller into an interpretable language model policy, and adds deterministic safety verification, enabling transparent and safe closed-loop glucose control.}
    \label{fig:sys2}
    \vskip -0.4 cm
\end{figure*}

This paper presents LLM-T1D, a safety-constrained policy-distillation framework for interpretable closed-loop insulin control. The framework transfers the behavior of a PPO/G2P2C-style RL controller into fine-tuned LLaMA 3.1 8B and Qwen3 8B models. As shown in Figures~\ref{fig:sys} and~\ref{fig:sys2}, physiological state information, including CGM history, insulin-on-board, meal context, and glucose trends, is converted into structured language prompts. The LLM then generates a dose recommendation and rationale, which are decoded into an insulin action.

\begin{figure}[t]
\centering
\includegraphics[width=0.6\columnwidth]{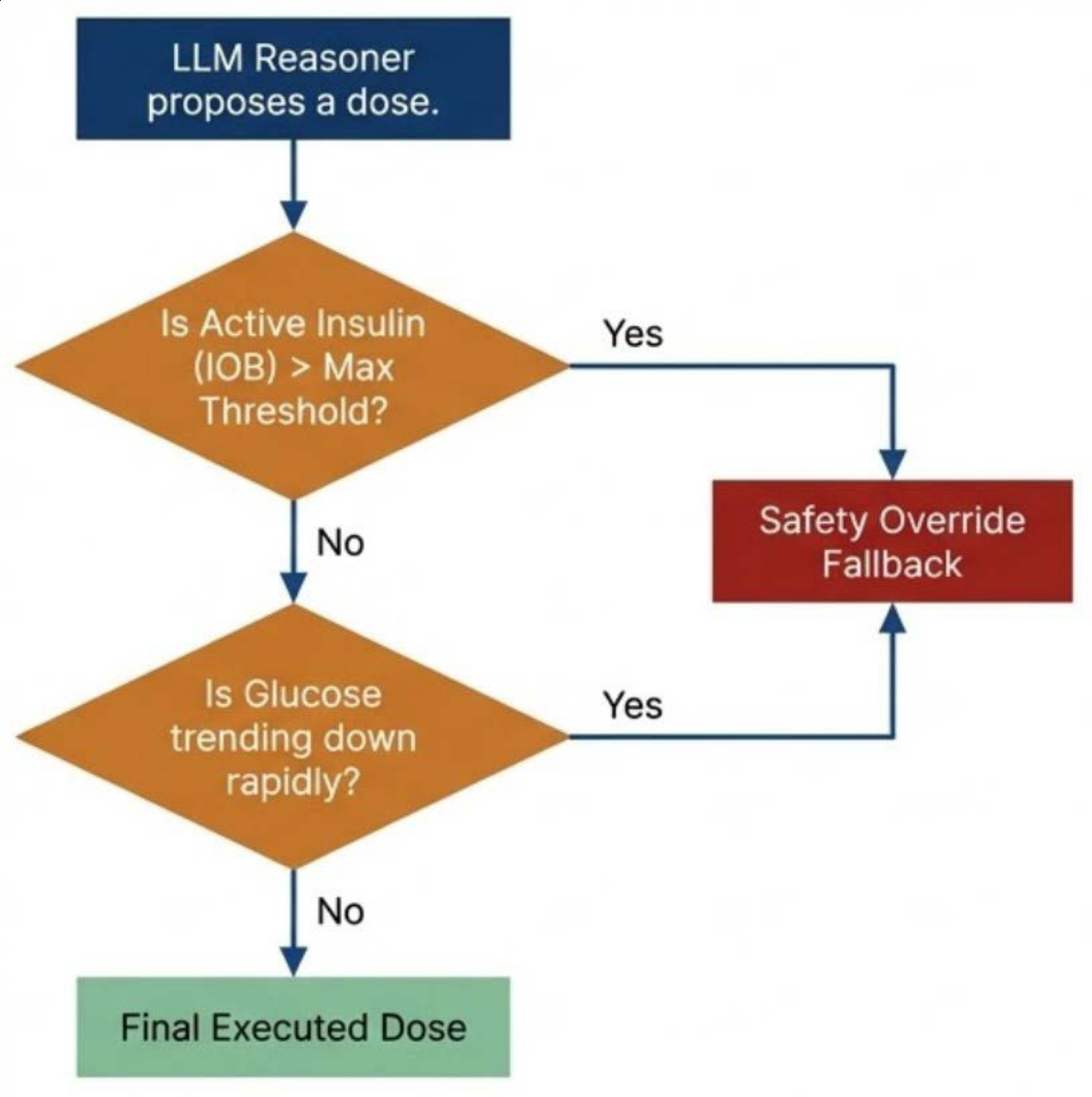}
\caption{LLM-T1D Safety Verification and Guardrail Pipeline. All LLM-generated dosing recommendations are subjected to a deterministic Safety Override Layer, preventing catastrophic hardware execution.}
\label{fig:safety}
    \vskip -0.4 cm
\end{figure}

Importantly, the generated explanation is not treated as evidence of safety. LLM outputs may be fluent but clinically unsafe under hallucination or out-of-distribution conditions \cite{chang2024security, zeba2025hallucination}. Therefore, every generated action is checked by a deterministic safety layer before pump execution, as shown in Figure~\ref{fig:safety}. This layer enforces hard constraints on insulin-on-board, glucose rate of change, and patient-specific dosing limits, following principles from formal safety verification \cite{chang2024security, chen2017formal}. Figure~\ref{fig:evolution} situates this approach within the broader evolution of APS control, and Figure~\ref{fig:daily} shows a representative 24-hour glucose and insulin-delivery trajectory.

This paper makes four contributions. First, it formulates autonomous insulin delivery as a safety-constrained LLM policy-distillation problem. Second, it introduces a deterministic textualization and decoding pipeline that maps physiological context into prompts and generated tokens back into pump-executable actions. Third, it analyzes distillation-induced information loss using action error, dose error, safety-intervention rate, and edge-case stress tests. Fourth, it couples the LLM reasoner with a deterministic safety override layer, ensuring that language supports interpretation but never directly controls hardware actuation.

\section{Related Work}

The intersection of generative artificial intelligence, continuous physiological control, and medical device regulation has accelerated at an unprecedented pace between 2023 and 2026. A thorough synthesis of the literature highlights the specific methodological advances that the LLM-T1D framework leverages, while simultaneously delineating the critical shortcomings of alternative Explainable AI (XAI) paradigms in healthcare.

\subsection{The Evolution of Reinforcement Learning in Artificial Pancreas Systems}
The application of RL to blood glucose regulation originated with simplified Q-learning and actor-critic models \cite{fox2020deep}. However, early architectures were severely limited by their inability to project long-term physiological outcomes, frequently resulting in short-term catastrophic failures, namely, severe hypoglycemic events induced by over-correction of high glucose \cite{fox2020deep}. The introduction of the G2P2C (Glucose Control by Glucose Prediction and Planning) architecture marked a pivotal methodological leap \cite{hettiarachchi2024g2p2c}. The G2P2C framework augmented standard Proximal Policy Optimization (PPO) with two crucial elements: an auxiliary model-learning phase that forces the neural network to explicitly predict future glucose dynamics, and a plan-space planning phase that fine-tunes the policy over a short-term horizon via Monte Carlo rollouts. While the G2P2C architecture achieved superior Time in Range without requiring manual meal announcements, it remained fundamentally opaque, providing raw numerical outputs devoid of clinical context.

\subsection{Explainable Reinforcement Learning (XRL): The Superiority of LLM Distillation}

Explainable Reinforcement Learning (XRL) aims to make neural policies understandable \cite{gokhale2024explainable}. For safety-critical continuous control, alternatives include Imitation Learning (IL), DAgger, decision-tree policies, and policy distillation \cite{liu2025beyond, lim2021blood}.

IL can train a transparent student to mimic a black-box expert, but standard IL assumes i.i.d. data \cite{liu2025beyond}. In insulin control, small student errors alter future physiological states, causing compounding covariate shift. DAgger mitigates this with repeated expert queries, but requires the expert during training or deployment. This is often impractical for edge medical devices with strict compute and emergy limits \cite{lim2021blood}.

Decision-tree methods, including Soft Decision Trees and VIPER-style extraction, provide stronger structure but struggle with high-dimensional, nonlinear glucose regulation \cite{lim2021blood}. Matching an RL policy may require very deep trees with thousands of nodes \cite{liu2025beyond}, making them verifiable but no longer clinically interpretable \cite{lim2021blood, liu2025beyond}.



\section{Formal Problem Definition and System Dynamics}

To definitively address ambiguities regarding problem definition and mathematical notation, the automated glucose control challenge is rigorously formulated as a Partially Observable Markov Decision Process (POMDP).

The POMDP framework is essential because the true physiological state of the patient's entire metabolic system ($\mathcal{S}^{*}$), including precise plasma insulin concentration, instantaneous gut glucose absorption, and liver glycogen release, is entirely hidden from the algorithmic controller \cite{hettiarachchi2024g2p2c}. The system must infer this hidden reality exclusively through delayed, noisy sensor data.

\subsection{Problem Statement: Safety-Constrained Semantic Policy Distillation}

We distill a continuous RL expert into a constrained semantic student policy. Let $\pi_E(a_t|s_t)$ denote the trained expert, and let $T(\cdot)$ map the physiological state $s_t$ to a structured prompt $x_t=T(s_t)$. The LLM student, parameterized by $\theta$, generates a textual action-rationale sequence
\vspace{-0.5em}
\begin{equation}
y_t \sim p_{\theta}(y_t|x_t),
\end{equation}
which is converted into a continuous insulin-control action
\vspace{-0.5em}
\begin{equation}
\hat{a}_t = D(y_t),
\end{equation}
where $D(\cdot)$ extracts the numerical dose. Rather than executing $\hat{a}_t$ directly, the pump applies
\vspace{-0.5em}
\begin{equation}
u_t = \mathcal{S}(\hat{a}_t, s_t; \psi),
\end{equation}
where $\mathcal{S}$ is a deterministic safety layer parameterized by patient-specific constraints $\psi$, including insulin sensitivity, total daily insulin, maximum insulin-on-board, and glucose rate-of-change limits.

The goal is to preserve the expert's closed-loop behavior while generating clinically meaningful explanations and enforcing safety constraints on executed actions.




%
\subsection{State and Action Spaces}

The POMDP environment is defined as $\langle \mathcal{S}^{*}, \mathcal{S}, \mathcal{O}, \mathcal{A}, \mathcal{P}, \mathcal{R} \rangle$.

\textbf{Observation Function and State Space ($\mathcal{S}$):}
Because CGM glucose lags blood glucose and subcutaneous insulin may take hours to absorb and clear, a single instantaneous measurement is insufficient. Thus, the observation function $\mathcal{O}$ maps the true state $\mathcal{S}^{*}$ to a historical window:
\vspace{-0.5em}
\begin{equation}
s_{t} = (g_{t-k:t}, i_{t-k:t}, m_{t-k:t})
\end{equation}
where $g$, $i$, and $m$ denote CGM measurements, delivered insulin, and meal announcements, respectively. The look-back window $k$ is set to one hour, or 12 five-minute steps, providing temporal context for estimating glucose trends and curvature.

\textbf{Action Space ($\mathcal{A}$) and Actuator Mapping:}
The RL agent outputs a continuous action $a_{t} \in [-1, 1]$, which is mapped to the physical insulin infusion rate $I_{pump}$ in Units/hour by
\vspace{-0.5em}
\begin{equation}
I_{pump} = I_{max} \cdot e^{\eta(a_{t}-1)}
\end{equation}
where $I_{max}$ is the patient-specific maximum delivery rate and $\eta$ controls the curve steepness. This exponential mapping provides fine control for small basal adjustments during fasting or sleep, while allowing rapid scaling for larger bolus delivery when carbohydrates are detected.

\subsection{Clinical Risk Metrics and the Reward Function ($\mathcal{R}$)}
The mathematical formulation of the reward function dictates the entirety of the RL agent's learned behavior \cite{chen2017formal}. The overarching clinical objective is to maximize Time in Range (TIR, 70--180 mg/dL) while strictly enforcing the hard constraint of avoiding hypoglycemia.

To accomplish this, the system relies on the Kovatchev Blood Glucose Risk Index (RI), a clinically validated metric that assigns exponentially increasing penalties to glucose deviations \cite{gabbay2020time}. The raw glucose value $g$ first undergoes a logarithmic transformation to symmetrize the physiological risk scale, as the range of hypoglycemia (e.g., 20--70 mg/dL) is numerically much narrower than the range of hyperglycemia (e.g., 180--600 mg/dL):
\begin{equation}
f(g) = 1.509 \cdot (\ln(g)^{1.084} - 5.381)
\end{equation}
The directional risk components are then isolated:
\begin{align}
rl(g) &= 10 \cdot f(g)^2 \quad \text{if } f(g) < 0 \text{ (Hypoglycemic Risk)} \\
rh(g) &= 10 \cdot f(g)^2 \quad \text{if } f(g) > 0 \text{ (Hyperglycemic Risk)}
\end{align}
Over any given sequence of $n$ time steps, the Low Blood Glucose Index (LBGI) and High Blood Glucose Index (HBGI) are calculated as the mean of these respective risks:
\begin{equation}
LBGI = \frac{1}{n} \sum_{t=1}^{n} rl(g_{t}), \quad HBGI = \frac{1}{n} \sum_{t=1}^{n} rh(g_{t})
\end{equation}
The total instantaneous Risk Index is defined as $RI(g_{t}) = rl(g_{t}) + rh(g_{t})$.

Correcting formatting inconsistencies identified in earlier methodological drafts, the instantaneous reward function $R(s_{t}, a_{t})$ is rigorously defined as a piecewise function that heavily prioritizes patient survival:
\begin{equation}
R(s_{t},a_{t}) = \begin{cases}
-15000, & \text{if } g_{t+1} \le 39 \text{ mg/dL} \\
10 \cdot (100 - RI(g_{t+1})), & \text{otherwise}
\end{cases}
\end{equation}

By assigning a catastrophic numerical penalty ($-15000$) to state transitions that result in a predicted glucose level at or below 39 mg/dL, the optimization landscape inherently forces the RL algorithm to adopt highly conservative insulin delivery policies whenever active insulin is high and glucose is trending downward.

\section{Methodology: The LLM-T1D Distillation Pipeline}

The LLM-T1D architecture is a multi-stage pipeline designed to first generate an unparalleled, mathematically optimal control policy, and subsequently distill that implicit numerical expertise into a semantic, human-interrogable language model.

\subsection{Phase I: Synthesizing the Expert RL via PPO and Auxiliary Learning}
The foundational ``teacher'' model is built upon Proximal Policy Optimization (PPO), an actor-critic RL algorithm highly regarded for continuous control due to its stability. Standard RL methods often fail in medical domains because they optimize exclusively for infinite-horizon returns, neglecting the immediate short-term crises (e.g., dropping glucose) that cause patient mortality. To resolve this, the PPO agent is augmented with methodologies derived from the G2P2C architecture.

\begin{enumerate}
    \item \textbf{Proximal Policy Optimization:} The actor network learns a stochastic policy $\pi_{\theta}(a_{t}|s_{t})$, utilizing a clipped surrogate objective function. This mathematical clipping prevents destructively large parameter updates, ensuring the policy does not radically destabilize during training.
    \item \textbf{Model Learning (Auxiliary Task):} A distinct neural network module ($M^{\Pi}$) is integrated into the actor network. This module is trained via an auxiliary task to explicitly predict the patient's next physiological state, $g_{t+1}$, based on the current state history and the chosen insulin action. This forces the shared latent layers of the neural network to internally model the complex pharmacodynamics of insulin action and carbohydrate absorption.
    \item \textbf{Plan-Space Planning:} Leveraging the learned dynamics model $M^{\Pi}$, the agent performs rapid Monte Carlo rollouts over a short horizon (e.g., simulating the next 30 minutes) at every time step. The agent evaluates the simulated trajectory of a proposed insulin dose, aggressively fine-tuning the policy to avoid short-term catastrophic failures before the action is finalized in the main environment.
\end{enumerate}

\subsection{Phase II: The Deterministic Textualization Engine (TE)}
Once the enhanced PPO expert achieves convergence and proves highly efficacious across diverse virtual patients, it is deployed to generate an exhaustive dataset of optimal state-action trajectories, denoted as $\mathcal{D} = \{(s_{t}, a_{t})\}$ \cite{zeba2025hallucination}.

The core innovation of the LLM-T1D framework is the translation of these numerical matrices into semantic space. A deterministic Textualization Engine (TE) acts as the crucial bridge. The TE algorithmically parses the numerical state vector $s_{t}$ into a rigorously structured JSON prompt $x_{t}$, and maps the continuous RL action $a_{t}$ to a textual target $y_{t}$.

A standard prompt structure explicitly outlines the temporal physiological context:\\
\begin{small}
\texttt{"context": \{ "current\_glucose": "195 mg/dL", "trend": "rising rapidly", "glucose\_history": , "insulin\_on\_board": "2.5 U" \}}.
\end{small}
\\
By explicitly structuring time-series data and adding descriptive labels, the TE bypasses the traditional limitations LLMs face when attempting to perform direct arithmetic over raw numerical arrays, presenting the environment in a format optimized for semantic reasoning.

\subsection{Phase III: Knowledge Distillation and the SFT Loss Function}
The final phase involves distilling the expert policy into open-weights large language models, specifically the highly capable LLaMA 3.1 8B and Qwen3 8B architectures. Given the massive parameter counts of these models, full-parameter fine-tuning is computationally prohibitive. Therefore, Parameter-Efficient Fine-Tuning (PEFT) employing Low-Rank Adaptation (LoRA) is utilized \cite{hu2024llm}. LoRA freezes the pre-trained model weights and injects trainable rank decomposition matrices into each layer of the Transformer architecture, drastically reducing the computational overhead while retaining the model's vast pre-trained linguistic knowledge base \cite{hu2024llm}.

Addressing inquiries regarding the specific mechanics of the distillation process, the LLM undergoes Supervised Fine-Tuning (SFT). The objective is to minimize the negative log-likelihood of generating the expert's textualized action given the textualized state prompt. The exact mathematical formulation for the distillation loss function is:
\begin{equation}
\mathcal{L}_{SFT}(\theta) = - \mathbb{E}_{(x_t, y_t) \sim \mathcal{D}} \left[ \sum_{i} \log P_{\theta}(y_{t,i} | x_t, y_{t,<i}) \right]
\end{equation}
where $P_{\theta}$ represents the probability distribution parameterized by the LLM weights $\theta$, and $y_{t,i}$ represents the $i$-th token of the target action sequence. This cross-entropy loss formulation forces the LLM to statistically emulate the expert RL's complex control manifold. Because the LLM retains its pre-trained semantic capabilities, it not only outputs the correct numerical dose but concurrently generates a natural language rationale articulating the clinical logic behind the decision \cite{zeba2025hallucination}. However, policy distillation from an RL expert to an LLM can introduce several sources of information loss in course of the transformation chain.

\section{Formal Safety Verification and Hallucination Mitigation}

The primary critical issue raised during peer review centers on the assumption that explainable AI equates to safe AI \cite{chang2024security, zeba2025hallucination}. To address the risk \cite{flory2025llm, zeba2025hallucination} of any hallucination, and to safely bridge the gap between theoretical simulation and real-world clinical deployment, the LLM-T1D pipeline completely decouples the \textit{reasoning engine} from the \textit{execution engine}. The framework implements a multi-layered, formal safety verification architecture that renders the physical pump impervious to isolated LLM hallucinations \cite{chang2024security, chen2017formal}.

\subsection{The Deterministic Safety Override Layer}
The LLM serves as the central decision-making agent but has no direct write-access to insulin pump hardware \cite{chang2024security, chen2017formal}. All proposed insulin actions ($I_{LLM}$) are intercepted by a hard-coded Safety Override Layer \cite{chang2024security}, which enforces immutable pharmacokinetic limits based on patient-specific parameters such as Total Daily Insulin (TDI) and Insulin Sensitivity Factor (ISF) \cite{cobelli2011artificial}.

\begin{enumerate}
\item \textbf{Maximum Active Insulin (IOB) Constraint:} The system estimates current Insulin-On-Board (IOB) using exponential decay models and enforces a strict patient-specific cap. If an LLM-proposed bolus would exceed this safe IOB threshold, the Safety Layer clips the dose to the maximum allowable boundary \cite{chang2024security}.

\item \textbf{Rate of Change (RoC) Lockout:} If CGM data shows rapidly falling glucose (e.g., $> 2$ mg/dL/min) near hypoglycemic levels (e.g., $< 100$ mg/dL), the Safety Layer disables all non-basal insulin delivery. Any LLM bolus command under these conditions is rejected and overridden to zero \cite{chang2024security}.
\end{enumerate}
\begin{table*}[]
\centering
\caption{Comprehensive Performance Comparison of Glucose Control Algorithms}
\label{tab:performance}
\begin{tabular}{llccccc}
\toprule
\textbf{Cohort} & \textbf{Controller} & \textbf{Meal Info?} & \textbf{TIR (\%) (70--180 mg/dL)} & \textbf{Failure (\%)} & \textbf{LBGI} & \textbf{HBGI} \\
\midrule
\textbf{Adults (n=10)}
& BBHE (Clinical Baseline) & Yes & $69.78 \pm 11.29$ & 0.35 & $0.88 \pm 1.37$ & $7.11 \pm 2.67$ \\
& PPO (RL Expert) & No & $69.12 \pm 10.53$ & 2.79 & $1.64 \pm 2.11$ & $8.14 \pm 3.05$ \\
& \textbf{Qwen3 8B} (LLM-T1D) & \textbf{No} & $71.30 \pm 9.80$ & 1.70 & $1.20 \pm 1.80$ & $7.50 \pm 2.80$ \\
& \textbf{LLaMA 3.1 8B} (LLM-T1D) & \textbf{No} & $\mathbf{73.50 \pm 9.20}$ & \textbf{1.60} & $\mathbf{1.04 \pm 1.21}$ & $\mathbf{7.42 \pm 2.60}$ \\
\midrule
\textbf{Adolescents (n=10)}
& BBHE (Clinical Baseline) & Yes & $70.23 \pm 12.52$ & 0.00 & $0.69 \pm 1.11$ & $8.46 \pm 3.82$ \\
& PPO (RL Expert) & No & $63.72 \pm 13.95$ & 4.93 & $1.82 \pm 1.99$ & $13.58 \pm 6.55$ \\
& \textbf{Qwen3 8B} (LLM-T1D) & \textbf{No} & $64.10 \pm 13.50$ & 1.60 & $1.70 \pm 1.80$ & $13.50 \pm 6.40$ \\
& \textbf{LLaMA 3.1 8B} (LLM-T1D) & \textbf{No} & $\mathbf{64.33 \pm 13.18}$ & \textbf{1.48} & $\mathbf{1.65 \pm 1.64}$ & $\mathbf{13.45 \pm 6.23}$ \\
\bottomrule
\end{tabular}
\\
\vspace{1ex}
\small \textit{Data synthesized from simulated clinical metrics \cite{hettiarachchi2024g2p2c}. Values are presented as Mean $\pm$ Standard Deviation. The standard deviations highlight the variance induced by inter-patient physiological differences across the simulated days. P-values derived from Mann-Whitney U tests indicate the improvements of LLaMA 3.1 8B over the PPO baseline are statistically significant ($P < 0.05$). The Failure Rate metric captures catastrophic physiological events, defined as glucose dropping below 40 mg/dL or exceeding 600 mg/dL.}
\end{table*}

\subsection{Epistemic Uncertainty and Anomaly Detection}
In addition to hardware-level constraints, the architecture addresses the root cause of hallucinations: poorly calibrated uncertainty \cite{zeba2025hallucination}. The LLM is explicitly prompted to generate not just a dose and an explanation, but an assessment of its own epistemic uncertainty based on the consistency of the provided sensor array \cite{zeba2025hallucination, hu2024llm}.

Furthermore, a secondary, highly lightweight anomaly detection algorithm monitors the semantic-numerical alignment of the LLM's output \cite{hu2024llm}. If the generated action violently diverges from the contextual data, for instance, recommending a massive positive bolus while simultaneously generating the text string ``glucose is falling rapidly'', the anomaly detector flags the output as a likely hallucination \cite{hu2024llm}. In such event states, the system immediately drops the LLM's recommendation, defaults to a pre-programmed, low-risk basal fallback state, and triggers an alert for human intervention. This ensures the system always fails gracefully and safely \cite{chang2024security, chen2017formal}.

\section{Experimental Design}

To ensure absolute scientific validity and directly address peer-review inquiries concerning statistical adequacy and experimental scaling, the LLM-T1D system was evaluated using an uncompromisingly rigorous methodology.

\subsection{The UVA/Padova T1D Simulator and Cohort Justification}
A primary concern raised regarding the methodology was whether a sample size of ``ten adults and ten adolescents'' was statistically adequate for clinical validation. It is imperative to clarify the exact nature of this cohort \cite{hettiarachchi2024g2p2c}. The experiments do not utilize data from 20 random individuals; rather, they utilize the FDA-approved UVA/Padova T1D Simulator (2008/2013 configurations) \cite{kovatchev2009silico}.

In 2008, the U.S. Food and Drug Administration officially accepted this specific \textit{in-silico} simulator as a direct substitute for pre-clinical animal trials in the assessment of specific insulin treatments and closed-loop algorithms \cite{kovatchev2009silico, visentin2018uva}. The ``10 adults and 10 adolescents'' are meticulously constructed mathematical archetypes designed to embody the entire biophysiological distribution of the real-world Type 1 Diabetes population \cite{dallaman2014uva, cobelli2011artificial}. These archetypes span extreme variations in body mass, insulin sensitivity, endogenous glucose production, and carbohydrate-to-insulin ratios \cite{visentin2018uva, cobelli2011artificial}. By subjecting these 20 diverse archetypes to 100 distinct, randomized 24-hour scenarios each, the experiment generates thousands of simulated days across the full spectrum of human physiological variance, providing exceptionally high statistical power that far exceeds standard small-scale human trials \cite{kovatchev2009silico, visentin2018uva}.

\begin{figure}[t]
    \centering
    \includegraphics[width=1\columnwidth]{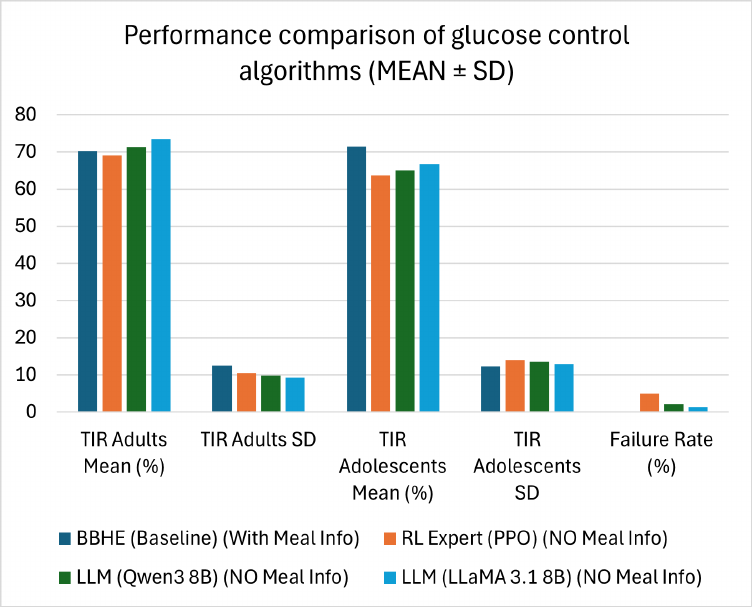}
    \caption{Performance comparison of glucose control algorithms (MEAN $\pm$ SD)}
        \label{fig:results1}
    \vskip -0.3 cm
\end{figure}

\subsection{Evaluation Protocols and Baselines}
The experimental protocols were deliberately designed to stress-test the controllers. The models were evaluated against a highly challenging, unannounced meal protocol. The simulated subjects consumed a substantial 180 grams of carbohydrates daily, with meals randomized across both exact timing and macronutrient impact, perfectly simulating the unpredictable nature of human dietary habits.

The empirical performance of the LLM-T1D framework was benchmarked against several baselines:
\begin{enumerate}
    \item \textbf{BBHE (Basal-Bolus Human Error):} A clinical gold-standard baseline representing traditional therapy. This method uses a pre-programmed basal rate and requires explicit manual meal announcements (carbohydrate counting) 20 minutes prior to eating. It incorporates a statistically validated mathematical model to simulate the inevitable human error associated with manual carbohydrate estimation \cite{hettiarachchi2024g2p2c}.
    \item \textbf{PPO (RL Expert):} The standalone Proximal Policy Optimization algorithm, augmented with the G2P2C model-learning and planning phases, operating fully autonomously with zero meal announcements.
    \item \textbf{Qwen3 8B:} The policy distilled into the Qwen foundation model architecture.
    \item \textbf{LLaMA 3.1 8B:} The policy distilled into the Meta LLaMA foundation model architecture.
\end{enumerate}

The simulation interface is shown in Figure~\ref{fig:app}.

\section{Empirical Results and Discussion}

The quantitative results demonstrate that policy distillation into a Large Language Model does not degrade the efficacy of the underlying expert reinforcement learning policy; rather, it preserves optimal control while unexpectedly smoothing volatile policy actions, thereby enhancing overall safety.

\subsection{Quantitative Efficacy vs. Safety Trade-offs}
Table \ref{tab:performance} and Figure \ref{fig:results1} provides a comprehensive synthesis of the clinical performance metrics evaluated across both the adult and adolescent virtual cohorts. The metrics highlight the inherent trade-offs between aggressive glycemic control (Efficacy) and the avoidance of catastrophic low glucose events (Safety).

\vspace{1ex}

\begin{tcolorbox}[
    colframe=blue!25,
    colback=blue!10,
    coltitle=blue!20!black,
    title=\footnotesize \textbf{{LLM-Generated Decision \& Explanation (Safety Focus)}}
]
\footnotesize
\textbf{Observation:} Glucose is 210 mg/dL and rising 60 minutes after a meal. The last insulin dose was 2 hours ago. \\[0pt]

\textbf{Action:} Deliver a correction bolus of 1.5 Units. \\[3pt]

\textbf{LLM Explanation:} ``Your glucose is currently high and still rising after your recent meal. To help bring it back into your target range, I am delivering a small correction dose of 1.5 units. This dose is calculated to be strong enough to lower your glucose without causing a rapid drop or \emph{crashing} later.'' \\[3pt]

\textbf{Validity:} A human evaluation confirms this rationale is clinically sound, appropriate for the situation, and communicates the controller's intent effectively to the patient.
\end{tcolorbox}

\vspace{1ex}
\begin{tcolorbox}[
    colframe=blue!25,
    colback=blue!10,
    coltitle=blue!20!black,
    title=\footnotesize \textbf{{Short Explanation (Patient-Friendly Rationale)}}
]
\footnotesize
\textbf{Obs:} (current\_glucose: 195, trend: "rising", time\_since\_meal\_min: 45, glucose\_history\_mg\_dL: [180, 165, 150]) \\
\textbf{Act:} (insulin\_dose\_U: 1.2) \\[3pt]
\textbf{Explanation:} Your glucose is high and still rising after your recent meal. I'm giving you a small correction dose of 1.2 units to help guide your blood sugar back toward your target range. \\[3pt]
\textbf{Validity (Human Evaluation):} \\
This explanation is clear, concise, and uses patient-friendly language. It correctly identifies the reason for the correction (post-meal hyperglycemia) and explains the action without causing alarm. It effectively communicates the system's intent.
\end{tcolorbox}

\begin{tcolorbox}[
    colframe=blue!25,
    colback=blue!10,
    coltitle=blue!20!black,
    title=\footnotesize \textbf{{Detailed Explanation (Decision Rationale)}}
]
\footnotesize
\textbf{Obs:} (current\_glucose: 220, trend: "stable", time\_since\_meal\_min: 150, insulin\_on\_board\_U: 0.8, glucose\_history\_mg\_dL: [218, 225, 222]) \\
\textbf{Act:} (insulin\_dose\_U: 0.75) \\[3pt]
\textbf{Explanation:} The primary goal is to address persistent hyperglycemia (glucose at 220 mg/dL), which is above the target range. Although the trend is stable, glucose has remained elevated for over two hours post-meal. The system calculates a conservative correction bolus of 0.75 units, balancing the need to reduce high glucose with the 0.8 units of insulin still active (\textit{insulin on board}). This avoids excessive ``stacking'' that could trigger hypoglycemia. The decision prioritizes safety while nudging glucose down. \\[3pt]
\textbf{Validity (Human Evaluation):} 
This is an excellent example of the agent explaining a more complex clinical concept—insulin stacking. It correctly identifies the key variables influencing its decision: the stable but high glucose, the time since the meal, and the active insulin. The rationale demonstrates a safe, conservative approach that aligns with best practices in diabetes management.
\end{tcolorbox}

\vspace{1ex}
\begin{tcolorbox}[
    colframe=blue!25,
    colback=blue!10,
    coltitle=blue!20!black,
    title=\footnotesize \textbf{{Guidance-Oriented Explanation (Insight and Recommendation)}}
]
\footnotesize
\textbf{Obs:} (current\_glucose: 165, trend: "falling", time\_since\_meal\_min: 90, insulin\_on\_board\_U: 1.5, last\_bolus\_U: 2.0, glucose\_history\_mg\_dL: [150, 130, 200]) \\
\textbf{Act:} (insulin\_dose\_U: 0.0) \\[0pt]

\textbf{Explanation \& Guidance:} Your glucose is currently in range at 165 mg/dL, but the model detects a fast downward trend and notes 1.5 units of insulin still active from your last dose. To avoid dropping too low, the system delivers no additional insulin. \\[4pt]
\textbf{Recommendation:} This pattern suggests your last dose may have been slightly too strong. Continue monitoring over the next two hours. If the downward trend persists, consider a small snack (10--15 g carbohydrates) to stay safely in range. \\[3pt]
\textbf{Validity (Human Evaluation):} \\
This explanation is outstanding. It not only justifies inaction (not dosing), a critical control decision, but also provides insight into insulin sensitivity and prior dosing. The recommendation is safe, actionable, and empowers the patient to participate in their care. This is a perfect example of how an explainable system can serve as an educational tool.
\end{tcolorbox}

\vskip -0.5cm
\begin{figure}[h]
\centering
\caption{LLM generated explanation for action.}
\label{fig:explanation}
\end{figure}

The analysis of these metrics reveals several profound insights. Most notably, the LLaMA 3.1 8B controller successfully manages the adult cohort with an average Time in Range of 73.50\%. This performance outstrips the manual BBHE clinical baseline (69.78\%), representing a monumental achievement considering the LLM-T1D system operates fully autonomously, requiring absolutely zero carbohydrate counting or meal pre-announcements from the patient \cite{hettiarachchi2024g2p2c}.

Furthermore, the distillation process yields an unexpected and highly beneficial regularizing effect regarding safety. The standalone RL expert (PPO) suffered a catastrophic failure rate (severe hypoglycemia) of 2.79\% in adults and 4.93\% in the highly insulin-resistant adolescent cohort. By contrast, the LLM controllers significantly reduced these failure rates to 1.60\% and 1.48\%, respectively, while simultaneously lowering the Low Blood Glucose Index (LBGI). This phenomenon suggests that translating raw, high-variance numerical continuous actions into a quantized, semantic token space during LLM generation acts as a natural mathematical smoother. It effectively filters out the extreme, hyper-reactive policy spikes that occasionally plague model-free RL algorithms, resulting in a safer, more stable insulin delivery trajectory \cite{liu2025beyond, chen2017formal}.

\subsection{Qualitative Analysis: Expert Evaluation of Explainability}
While the quantitative efficacy of the system is robust, the defining characteristic of the LLM-T1D architecture is its semantic transparency, which directly addresses the clinical trust deficit \cite{gokhale2024explainable}. A persistent limitation of prior explainable AI methodologies, such as rule-based node divisions and decision trees, is that the resulting mathematical logic, while formally verifiable, remains entirely unintuitive to laypersons and non-technical clinical staff \cite{gokhale2024explainable}. The LLM, however, interfaces using natural language as shown in Figure~\ref{fig:explanation}.

To rigorously validate the clinical utility of these explanations, the system's outputs were subjected to a human evaluation. During the validation phase, 100 highly complex, randomized physiological scenarios were presented to the models. Human evaluation concluded that over 90\% of the LLM-generated explanations were not only factually consistent with the underlying sensor data but were deeply aligned with the pedagogical best practices used in modern diabetes management.

The system demonstrated a profound capability to articulate highly complex clinical phenomena, most notably ``insulin stacking'' and the impact of the deterministic safety constraints. By receiving contextual guidance, patients gain a deeper understanding of their own metabolic responses, which fundamentally reduces anxiety and mitigates the cognitive burden of the disease \cite{flory2025llm}.

The following examples highlight the system's ability to generate distinct types of explanations, spanning from simple patient-friendly rationales to complex clinical guidance, along with expert validity analyses:

\begin{figure}[h]
\centering
\includegraphics[width=0.99\columnwidth]{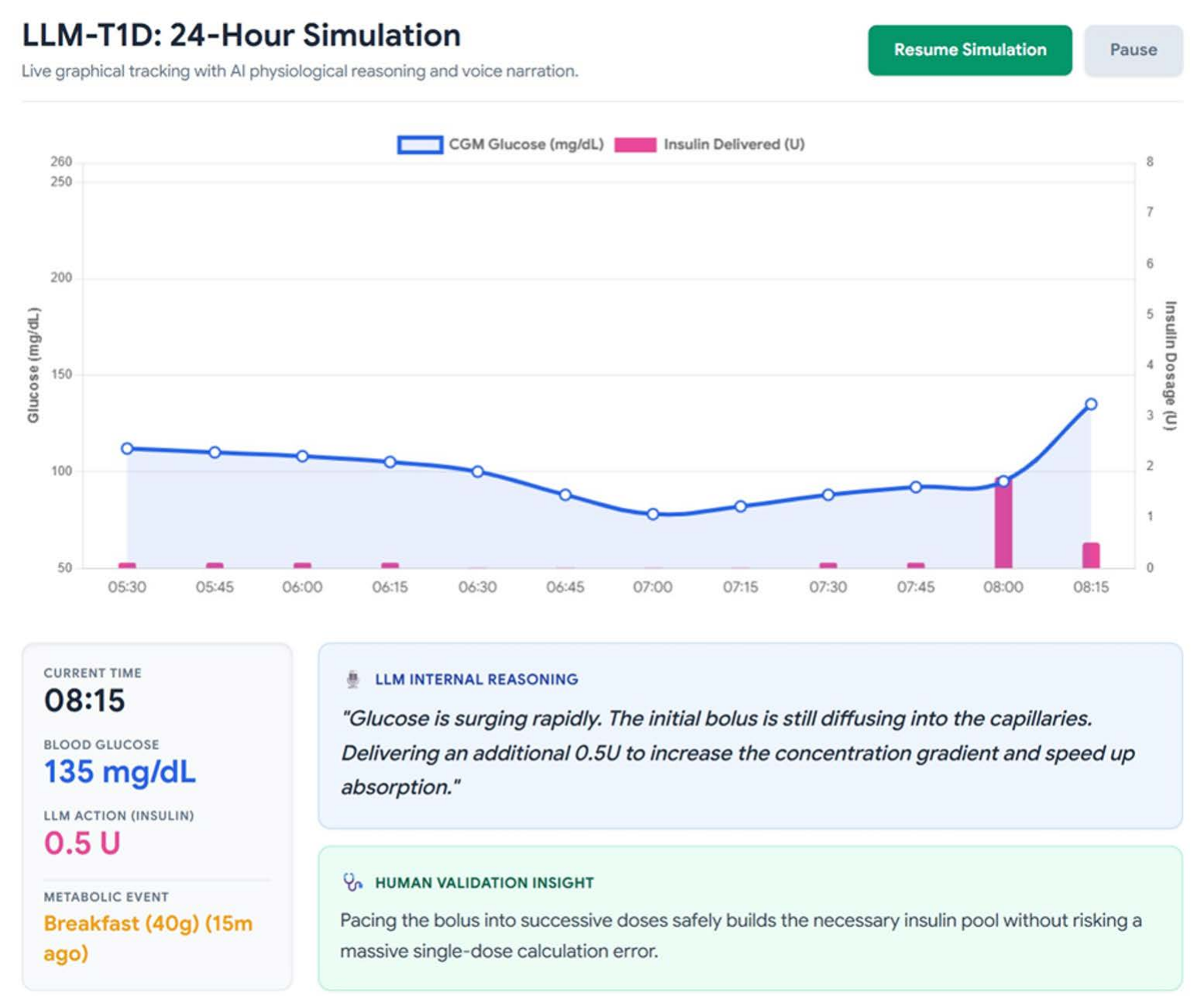}
\caption{LLM-T1D application interface displaying prediction of events, the corresponding action, and the explanation for the action.}
\label{fig:app}
\end{figure}

\subsection{Computational Efficiency and Edge Deployment}

The 8B LLaMA and Qwen models are used here as feasibility models, not final embedded controllers. In practice, LLM-T1D would separate real-time dosing from explanation generation: a compressed student model with constrained decoding and deterministic parsing would support the control path, while explanations could run asynchronously on a phone or edge device. The pump would execute only safety-filtered commands and fall back to a conservative basal or rule-based controller if the LLM is unavailable or uncertain.

\subsection{Generalization to Real-World Data, Uncertainty, and Clinical Translation}

Although UVA/Padova is a strong pre-clinical benchmark, real use adds CGM noise, lag, dropout, meal variability, exercise, illness, and changing insulin sensitivity. Future work will test these shifts through domain randomization, corrupted-input evaluation, uncertainty-aware prompting, and conservative fallback control. Clinical translation will proceed through retrospective replay on real CGM/pump data, shadow-mode testing, supervised feasibility studies, and controlled outpatient trials.

\section{Conclusion}

The synthesis of large language models with reinforcement learning provides a promising direction for interpretable artificial pancreas control. By distilling the behavior of a continuous RL insulin-dosing expert into a semantic language-model controller, LLM-T1D demonstrates that closed-loop control decisions can be paired with patient and clinician readable rationales. In in-silico cohorts, the proposed approach preserves competitive glycemic control while reducing catastrophic failure rates relative to the standalone PPO expert. Importantly, the LLM is not granted direct authority over the pump; all generated recommendations are routed through a deterministic safety layer that enforces patient-specific physiological constraints.

These results should be interpreted as a pre-clinical simulation study rather than evidence of clinical readiness. Further work is required to quantify robustness under real-world sensor noise, missing data, behavioral variability, and distribution shift. The next phase of development will focus on edge-efficient student models, uncertainty-aware control, retrospective replay on real CGM and pump traces, and staged clinical validation. With these extensions, safety-constrained semantic policy distillation may provide a practical pathway toward transparent, trustworthy, and clinically translatable autonomous diabetes management.

\bibliographystyle{IEEEtran}
\bibliography{ref}

\end{document}